# LUKE-Graph: A Transformer-based Approach With Gated Relational Graph Attention For Cloze-style Reading Comprehension

**Shima Foolad**
Department of Electrical & Computer Engineering
Semnan University
Semnan, Iran
sh.foolad@semnan.ac.ir

**Kourosh Kiani**[*]
Department of Electrical & Computer Engineering
Semnan University
Semnan, Iran
kourosh.kiani@semnan.ac.ir

March 12, 2023

## Abstract

Incorporating prior knowledge can improve existing pre-training models in cloze-style machine reading and has become a new trend in recent studies. Notably, most of the existing models have integrated external knowledge graphs (KG) and transformer-based models, such as BERT into a unified data structure. However, selecting the most relevant ambiguous entities in KG and extracting the best subgraph remains a challenge. In this paper, we propose the LUKE-Graph, a model that builds a heterogeneous graph based on the intuitive relationships between entities in a document without using any external KG. We then use a Relational Graph Attention (RGAT) network to fuse the graph's reasoning information and the contextual representation encoded by the pre-trained LUKE model. In this way, we can take advantage of LUKE, to derive an entity-aware representation; and a graph model - to exploit relation-aware representation. Moreover, we propose Gated-RGAT by augmenting RGAT with a gating mechanism that regulates the question information for the graph convolution operation. This is very similar to human reasoning processing because they always choose the best entity candidate based on the question information. Experimental results demonstrate that the LUKE-Graph achieves state-of-the-art performance on the ReCoRD dataset with commonsense reasoning.

## 1 Introduction

A common way to evaluate the capability of a computer's language understanding is the machine reading comprehension (MRC) task in natural language processing (NLP). To accomplish this task, the researchers have developed artificial intelligence models that can automatically answer the questions given in a document. Applying the transformer-based pre-trained models, such as BERT [1] and RoBERTa [2], has dramatically improved the performance of these models in public datasets. The intuition behind pre-trained models is that first understand the language and then be able to use it to do any downstream task in that language. Using the pre-trained models is very effective for supervised learning tasks such as MRC as there is not enough annotated data for training.

Despite the tremendous success of the pre-trained models in MRC [1–5], there is still a real challenge to bridge the gap between the human performance and the models in datasets with deep commonsense reasoning, e.g., ReCoRD [6]. Since little supervised information is available during pre-training, such models do not have explicit knowledge of inference concepts. Therefore, the representations from these models fall short in their ability to support reasoning.

---
[*]Corresponding author

Often, entities play an important role in such datasets, and, there is reasoning about the relationships between them. Hence, Yamada et al. [7] have developed a language understanding model with knowledge-based embeddings called LUKE. They have extended RoBERTa by appending entities to the input of the transformer model and considering them as independent tokens. It will enable LUKE to utilize an entity-aware self-attention mechanism to explicitly capture the connections between entities [8]. Although LUKE has archived impressive performance compared other transformer-based models, it is still not enough to distinguish between entities with the same strings and those that are unrelated. In other words, it does not take into account the intuitive relationships between entities in the long text.

Most of the recent works [9–12] have integrated external knowledge sources (e.g., WordNet and NELL) and BERT-large as a base model for handling commonsense reasoning in MRC. Therefore, they inject general knowledge graphs (KG) and pre-trained entity embeddings explicitly. This limits their ability to select the most relevant mentioned entities in KG, especially for the ambiguous words that, if not selected correctly, can create knowledge noise. Besides, choosing the best sub-graphs from external knowledge and using its information remains a challenge.

To address the above challenge, in this work, we propose a transformer-based approach called LUKE-Graph, which uses pre-trained LUKE as a base model to output an entity-aware representation for each word and entity token in a document. Following BAG [13], we also transform all entities in the document into a graph where the nodes correspond to the entities and edges represent the relationships between them. We then import the entity representations from LUKE into a Relational Graph Attention Network named RGAT [14] to learn a relationship-aware representation of entities. In this way, we solve the aforementioned problem of intuitively relating entities by superimposing the LUKE and RGAT layers. Moreover, similar to [15], we extend RGAT with a gating mechanism that control the question information used in the graph convolution operation. The experimental results demonstrate that the LUKE-Graph achieves excellent performance on MRC with commonsense reasoning. We can summarize our contributions as:

- We construct a heterogeneous graph based on the intuitive relationship between entities in a document as well as the reasoning on them without using any external KG.

- We choose the LUKE model among transformer-based models such as BERT, RoBERTa, XLNET, etc. Because it can explicitly capture the relationships between entities by treating them as independent tokens.

- We enrich the learned entity representation of the pre-trained LUKE model with RGAT, which proved valuable for ReCoRD dataset with commonsense reasoning.

- We enhance RGAT to Gated-RGAT by incorporating the question information during reasoning through a question-aware gating mechanism that is very similar to human reasoning processing.

## 2   Related Work

In this section, we briefly review recent works from three perspectives:

**Transformer-based Models:** In recent years, the development of transformer architecture has created a revolution in NLP tasks by offering accuracy comparable to human performance or even higher. Hence, it becomes a basic architecture in the pre-trained language models such as BERT, RoBERTa, and XLNET [3]. Due to the promising results of the pre-trained models on downstream tasks, researchers [4,16–19] have extended them by defining some forms of sparse attention pattern to adopt the model for long documents. Longformer [4] has introduced a fixed-size window of attention surrounding each token to reduce computation. Since some tokens are critical for learning task-specific representations, e.g., query tokens in MRC, Longformer adds global attention to a few pre-selected input tokens. Similar to full attention [1,2], the global tokens attend to all the input tokens, and all tokens in the input sequence attend to it. Ainslie et al. [18] named Extended Transformers Construction (ETC), which is closely related to Longformer, defining some new extra tokens as global that do not correspond to any input tokens. Zaheer et al. [19] built the BigBird model on the work of ETC, adding random connections between inputs to this structure. Also, Jia et al. [20] extracted keywords from the context to direct the model's focus to crucial details. However, the keywords don't make any sense together. Recently, researchers have proposed methods such as ERNIE [10], Know-BERT [21], and LUKE, which treat entities as independent tokens besides input words. ERNIE and Know-BERT enhanced contextualized word representations by learning static entity embeddings from an external knowledge source. Meanwhile, the LUKE model applies a new pretraining task to learn entity representations and it trains using a substantial corpus of entities annotated data. It also improves the transformer architecture using an entity-aware self-attention mechanism. As shown in recent work, the injection of prior knowledge information can significantly improve the existing pre-training models. Accordingly, we apply the LUKE pre-trained model to extract input features.

**Graph-based Models:** It has proven graphs to be an effective way to represent complex relationships between objects and extract relational information [22]. Recently, researchers [13,23–25] have studied graph-based models for MRC with commonsense reasoning. Entity-GCN [23] and BAG used Relational Graph Convolutional Networks (R-GCN) to comprehend the relationships of entities in text. Both of them built document-based entity graphs by matching entity candidate strings in the text and using them as nodes in the graph. The models represent nodes without fine-tuning by relying on token-level (GLoVe) and context-level (ELMo) features. Inspired by CFC [25], the HDEGraph model [24] applied both self-attention and co-attention to learn entity node representations. Besides, some studies regard the question context for the graph because it also provides crucial information that shouldn't be lost. Tang et al. [15] developed a path-based reasoning graph and used a gating mechanism to add the question context to RGCN. While Wu et al., [24] joined the subgraph with the entire question as a node and dynamically expanded subgraphs to efficiently utilize the question information and KB. These studies relate to our work because we also use the extended RGCN to capture reasoning chains between entities in a document and add the question information to the graph. Unlike all mentioned models that use a static entity representation, we use transformer-based models to obtain the abstract feature representations for each node.

**Combined Graph and Transformer Models:** As mentioned, transformer-based models have achieved striking success in a wide range of NLP tasks. Also, GCN is the most effective architecture and has led to state-of-the-art performance on MRC tasks. Hence, a considerable number of studies such as Graph-BERT [27] and Graph Transformer [28] have generalized the transformer architecture on arbitrary graphs to take advantage of both. They first created subgraphs or original graphs from a given document and then passed the nodes to the transformer layer instead of words in the document. Therefore, they do not consider the information interaction over words in long documents. While our work incorporates the enriched representations of interaction between words of the document using a transformer-based model in the graph reasoning process. Some studies such as SKG-BERT [11], KT-NET [12], and KELM [9] utilize an external KG such as WordNet, ConceptNet and NELL and BERT-large as a base model. While we do not use any additional KG and only transform the document entities into a graph to consider the relationships between them.

## 3 Methods

In this section, we describe our method and show its framework in Fig. 1. The LUKE-Graph method consists of two main components, namely a transformer-based module to extract the entity-aware representation, and a graph-based module to use the relation-aware representation. In the transformer-based module, we apply a pre-trained language model called LUKE that treats the words and entities in a given document as independent input tokens and outputs contextualized representations using an entity-aware self-attention mechanism. In the graph-based module, we construct a heterogeneous graph that relates the entities within a sentence and the same entities in different sentences. We then import the contextualized entity representations from the LUKE into two-layer RGAT to resolve the relationships between entities in the document before answering the questions. Moreover, we optimize each RGAT layer with a question-based gating mechanism and call it Gated-RGAT. Finally, we compute a score for each candidate entity using a linear classifier and select the candidate with the highest score as the final answer.

### 3.1 Transformer-based Module

**Embedding Layer:** Since using pre-trained language models is one of the most exciting directions for any task in NLP, we use the transformer-based model of LUKE. The model takes the following input sequence: [CLS] token, the tokenized questions (q1, q2, …, qn) along with a [PLC] special token in the position of missing entity (placeholder), two [SEP] tokens as a separator, the tokenized document (d1, d2, …, dm), [SEP] token. Further, it takes [MASK] token for the question placeholder and each candidate entity appearing in the document. Briefly, the form of input sequence is *"[CLS]+question+[SEP]+[SEP]+document+[SEP]+entities"*. We compute the input embedding vector of each token (word or entity) as the summation of the following three embeddings: token embeddings, position embeddings, and segment embeddings [1]. A token embedding is a numerical representation of the corresponding token learned during pretraining the LUKE model. Following RoBERTa, we generate word position embeddings based on the positions of the tokens in a sequence that provides a sense of order to the input data. While we average the position embeddings of the corresponding word tokens to derive entity position embeddings, as illustrated in Fig. 2. The segment embeddings represent the type of token, whether it is a word or an entity.

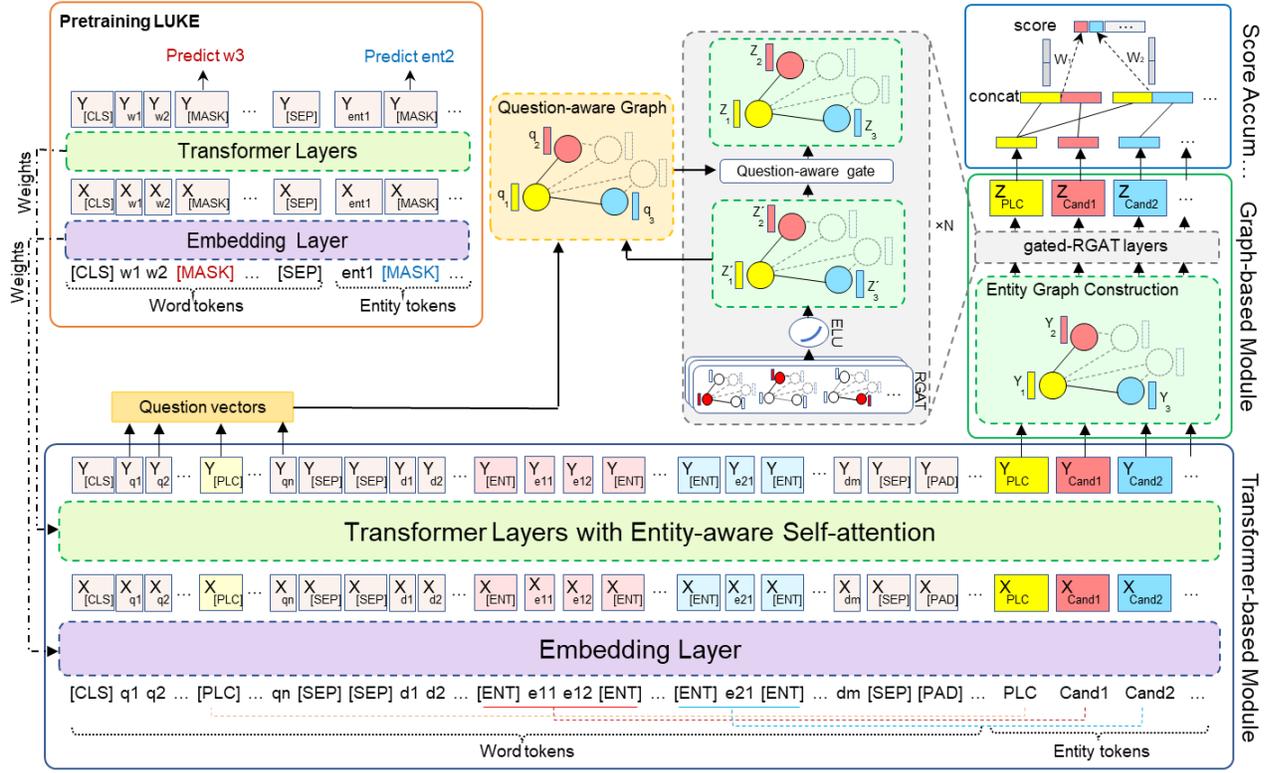

**Fig. 1.** The framework of the LUKE-Graph method.

**Transformer Layers with entity-aware Self-attention:** The main part of the transformer [29] is the self-attention mechanism that compares all input tokens by an attention score between each pair of them. Word and entity tokens alike undergo self-attention computations after layer embedding. It comprises three matrices: query, key, and value by multiplying the token embedding by three weight matrices that are trained during the training process. The only difference between the LUKE attention mechanism and the attention used in the original Transformer [29] is that a different query matrix is used by the LUKE based on token types. Formally, given a sequence of token embeddings $X_1, X_2, ..., X_P$, where $X_i \in \mathbb{R}^L$, we compute the output contextualized embeddings $Y_1, Y_2, ..., Y_P$, where $Y_i \in \mathbb{R}^H$ using the weighted sum of the input vectors transformed by the value matrix.

$$Y_i = \sum_{j=1}^{P} a_{ij} V X_i$$

$$a_{ij} = \mathrm{softmax}(\frac{QX_i KX_j^T}{\sqrt{H}}) \quad where\ Q = \begin{cases} Q_{w2w} & if\ X_i \in w, X_j \in w \\ Q_{w2e} & if\ X_i \in w, X_j \in e \\ Q_{e2w} & if\ X_i \in e, X_j \in w \\ Q_{e2e} & if\ X_i \in e, X_j \in e \end{cases} \qquad (1)$$

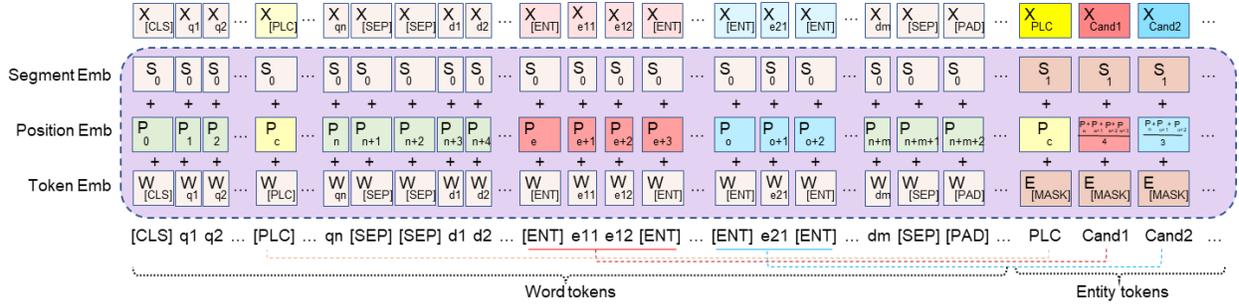

**Fig. 2.** Embedding layer.

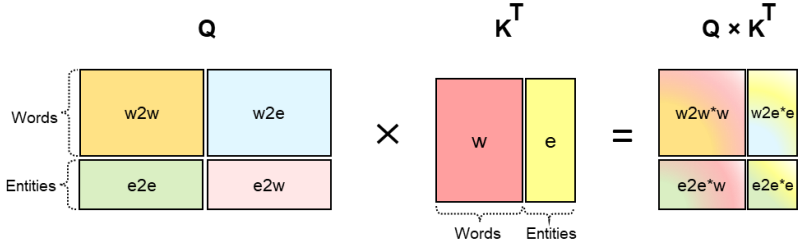

**Fig. 3.** Visualizing the multiplication of Q in K in entity-aware self-attention mechanism.

Where $Q$, $K$, $V \in \mathbb{R}^{H \times L}$ represent the query, key, and value matrices, respectively. To calculate the attention score $(a_{ij})$ among types of tokens, words $(w)$, or entities $(e)$, there are four query matrices. P and H denote the length of the input sequence, and the dimension of hidden states, respectively. For better understanding, we visualize the multiplication of Q in K in Fig. 3.

### 3.2 Graph-based Module

**Entity Graph Construction:** let an entity graph be denoted as G = {V, E}, where V indicates node representations and E is the edges between nodes. All the entities specified in the document along with the missing entity of question are used as nodes. Undirected edges are determined between the pairs of nodes that are located in the same sentence (SENT-BASED edge) and those with the same entity string in different sentences (MATCH edge). Furthermore, the node corresponding to the missing entity of the question is connected to all other nodes (PLC edge). See Fig. 4 for an illustration of graph construction.

Inspired by the BAG model [13], we build a graph with two differences as follows:

1. Defining the relationship between entities within and across sentences instead of paragraphs. Due to the paragraph texts being long, most of the entities are incorrectly connected in the BAG model. While generally, the entities that appear in a sentence are related to each other.
2. Assigning a node for the missing entity of the question (placeholder) and its relationship with all nodes in the graph, which is ignored in the BAG model. As proven in experiments, the placeholder node plays a significant role in the graph (as shown in Section 4.4), and important information for commonsense reasoning is lost without it.

**Gated-RGAT Layers:** To aggregate the information between the entities on a graph structure, we used one of the most popular GCN architectures named RGAT [14]. It extends the attention mechanisms of the GAT layer in Veličković et al. [30] to the relational graph domain from Schlichtkrull et al [31]. For this purpose, we import the entity representations of the LUKE model into the RGAT layers. For each layer, we define the $i$-th node by a representation of $Y_i \in \mathbb{R}^L$ as an input and an updated representation of $Z_i \in \mathbb{R}^{L'}$ as output. Moreover, the edges as mentioned above and their relation types (MATCH, SENT-BASED, or PLC) are considered inputs. In an RGAT layer, an attention score is computed for every edge $(i, j)$ and relation type $r$ named $a_{ij}^{(r)}$, which indicates the importance of the features of the neighbor $j$ to the node $i$, i.e.

**Document: Puerto Rico** on Sunday overwhelmingly voted for statehood. But **Congress**, the only body that can approve new states, will ultimately decide whether the status of the **US** commonwealth changes. Ninety-seven percent of the votes in …, official results from the **State Electoral Commission** show. Today, we the people of **Puerto Rico** are sending a strong and clear message to the **US Congress** ... and to the world ... claiming our equal rights as **American** citizens, **Puerto Rico** Gov. **Ricardo Rossello** said in a news release and **Puerto Rico** voted Sunday in favor of statehood

**Question:** For one, they can truthfully say, "Don't blame me, I didn't vote for them," when discussing the [PLC] presidency.

(a)

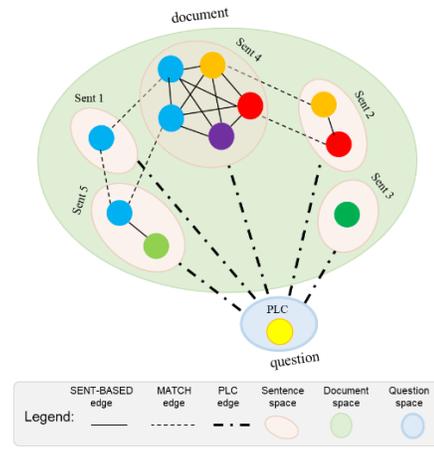

(b)

**Fig. 4.** the right side (a) indicates a constructed heterogeneous graph for the left side sample (b). Each ellipse in the document ellipse illustrates nodes belonging to a sentence. The same-colored nodes denote that they correspond to an identical entity. While the normal dashed lines connecting the node pairs correspond to MATCH edges. The solid lines connecting the node pairs correspond to SENT-BASED edges. The dash-dot lines connect the placeholder node (node corresponding to the missing entity of the question that is yellow in the graph) and all other nodes correspond to PLC edges. Note, we ignore the connection of the placeholder node with all other nodes for good visualization.

$$\begin{aligned} h_i^{(r)} &= W^{(r)} Y_i \\ e_{ij}^{(r)} &= \text{LeakyReLU}(h_i^{(r)}.Q^{(r)} + h_j^{(r)}.K^{(r)}) \\ a_{ij}^{(r)} &= \frac{\exp(e_{ij}^{(r)})}{\sum_{r' \in \Re} \sum_{k \in N_{r'}(i)} \exp(e_{ik}^{r'})} \end{aligned} \quad (2)$$

Where $W^{(r)} \in \mathbb{R}^{L \times L'}$, $Q^{(r)} \in \mathbb{R}^{L' \times D}$, and $K^{(r)} \in \mathbb{R}^{L' \times D}$ are the learnable parameters for linear transformation, query, and key attention kernels, respectively. $\Re$ represents the set of relations types and $N(i)$ denotes the number of nodes connected to $i$. Once acquired, the aggregation step is combined with the normalized attention scores to determine an updated representation for each node as follows:

$$Z_i' = \sigma(\sum_{r \in \Re} \sum_{j \in N(i)} a_{ij}^{(r)} h_j^{(r)}) \quad (3)$$

Where $\sigma$ is a nonlinearity (we use ELU). Similar to [29], RGAT uses multi-head attention to strengthen self-attention. In particular, Equation 3 is applied to the K independent attention mechanisms, which are then concatenated to produce the output representation as follows:

$$Z_i' = \bigoplus_{k=1}^{K} \sigma(\sum_{r \in \Re} \sum_{j \in N(i)} a_{ij}^{(r,k)} h_j^{(r,k)}) \quad (4)$$

Where $\oplus$ denotes concatenation, $a_{ij}^{(r,k)}$ are the normalized attention scores under relation r and computed by the $k$-th attention mechanism, and $h_i^{(r,k)} = W^{(r,k)} Y_i$. Since the use of multi-head attention creates of $KL'$ features instead of $L'$ for each node, we use averaging instead of concatenation in the last layer of RGAT.

Inspired by [15], we add a question-aware gating mechanism to RGAT that adjust the update message based on the question. This gating mechanism, like human reasoning processing, considers the question when choosing the entity information. Fig. 5 illustrates the question-aware gate for each node. For this purpose, first we create a question representation ($q_i$) for every node $i$ by a weighted sum of the question vectors encoded by LUKE, i.e.

$$q_i = \sum_{j=1}^{n} w_{ij} Y_{qj}$$

$$w_{ij} = \sigma(f_g([Z'_i; Y_{qj}])) \quad (5)$$

Where $Y_{qj}$ denotes the representation of $j$-th question token encoded by LUKE, σ is the sigmoid function, $[Z'_i; Y_{qj}]$ is a concatenation of $Z'_i$ and $Y_{qj}$, and $f_g(.)$ represents a single-layer multilayer perceptron (MLP). Finally, we update the message of $Z'_i$ based on the question representation as follows:

$$\alpha_i = \sigma(f_s([Z'_i; q_i]))$$

$$Z_i = \alpha_i \odot \tanh(q_i) + (1-\alpha_i) \odot Z'_i \quad (6)$$

Where $f_g(.)$ is a single-layer MLP, σ denotes the sigmoid function, and $\odot$ is element-wise multiplication.

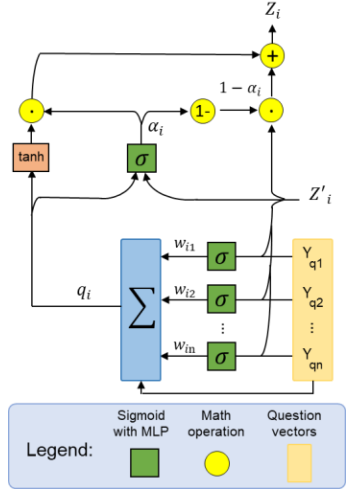

**Fig. 5.** Question-aware gate for node $i$.

### 3.3 Score Accumulation

We concatenate the final representation of the placeholder (PLC) with each candidate entity and then use a linear classifier to compute the candidate score. We use binary cross-entropy loss averaged across all candidates to train the model, and we select the candidate with the highest score as the final answer $e^*$:

$$e^* = \underset{e \in E}{\arg\max}\, f_o([Z_{PLC}; Z_e]) \quad (7)$$

Where $E$ denotes the set of entity candidates, $f_o(.)$ is a linear classifier, and $[Z_{PLC}; Z_e]$ is a concatenation of the PLC representation ($Z_{PLC}$) and the candidate representation ($Z_e$).

## 4 Experiments

In this section, we validate the effectiveness of the proposed method on a cloze-style reading comprehension dataset and compare it with other state-of-the-art methods.

### 4.1 Implementation Details

Our transformer-based module configuration follows the LUKE model. The hyper-parameters used in LUKE are shown in Table 1. The module structure follows the large LUKE model, which includes 24 hidden layers, 1024 hidden dimensions ($L = 1024$), 64 attention head dimensions ($H = 64$), and 16 self-attention heads. Therefore, a word token embedding has 1024 dimensions, while an entity token embedding has 256 dimensions. With a dense layer, the entity token embedding dimension is converted from 256 to 1024. We tokenize the input text using RoBERTa's tokenizer, which has a vocabulary of 50K words. Also, there are 500K common entities in the entity vocabulary along with two special entities, namely [MASK] and [UNK]. When an entity is missing from the vocabulary, the [UNK] entity is used in its place. Each entity's token embedding is initialized during finetuning using the [MASK] entity. Note the [MASK] word for the masked language model and the [MASK] entity for the masked entities in the pretraining task are the two [MASK] tokens that the LUKE employs. When lifting weights from pre-trained LUKE, the original query matrix $Q$ serves as the initialization matrix for the query matrices ($Q_{w2w}, Q_{w2e}, Q_{e2w}, Q_{e2e}$) in the self-attention mechanism. In our graph-based module, we randomly initialize a two-layer RGAT along with the ELU activation function. The first layer has 8 attention heads with 1024 hidden state size, while the second one has 1 head with 1024 hidden size ($L' = 1024$). Also, the number of relation types in our experiment is set to $\Re = 3$ and the dimensionality of the query and key are both D = 1. We implemented the graph-based module using the PyTorch Geometric library[1] [32]. We use AdamW to optimize the model based on performance on the development set.

Since we use the LUKE model as a baseline, we apply our method to its official repository[2]. We run our method using an Amazon EC2 p3.8xlarge instance with four GPUs of Tesla V100 SXM2 16 GB. A single model trained with 2 batch sizes and 2 epochs takes about 2 hours on the ReCoRD dataset. While the LUKE paper conducted the experiments on a server with eight V100 GPUs and two Intel Xeon E5-2698 v4 CPUs. Table 2 shows the different configurations between ours and LUKE. For a fair comparison, we run the official LUKE repository without any change on our server and achieved a lower development score compared to the reported one in the paper (See Table 2, Dev score column). Hence, we compare the LUKE model using the results obtained by the official LUKE code in the following.

### 4.2 Dataset

Our experiments are conducted on a cloze-style reading comprehension dataset named ReCoRD [6]. It is a challenging dataset that requires commonsense reasoning to answer the cloze-style questions. ReCoRD contains 120k automatically generated questions and documents from 70,000 CNN and Daily Mail news articles which have been validated by experts. Each cloze-style question is formulated as a sentence with the missing entity named placeholder. Therefore, an MRC system is expected to find a correct entity as an answer. In other words, the appropriate answer is chosen from among all entities in the related document to fill the placeholder in the question. Formally, given a document $D$ describing an event, a cloze-style question $Q([PLC])$ with a missing entity indicated by $[PLC]$, and a set of entity candidates $E$ marked in $D$, an MRC system is expected to choose an entity $e \in E$ that best fits $[PLC]$, i.e.,

$$e^* = \underset{e \in E}{\mathrm{argmax}}\, p(Q(e) | D) \tag{8}$$

Where $\mathrm{argmax}\, p(.)$ determines the most probable entity as $e^*$. Fig. 4(a) shows a ReCoRD example which $E$ are indicated by bold color text in the document.

There are about 100k samples in the training set, 10k samples in the development set, and 10k samples in the test set on ReCoRD. Since in this dataset, the test set is not publicly released, we compare models on the development set. Models are evaluated on the basis of the Exact Match (EM) and F1 criteria. The EM criterion is a binary number that indicates whether the answer produced by the model exactly matches the correct answer. The F1 measure calculates the amount of word overlap between the answer generated by the model and the actual correct answer. Human performance in ReCoRD has reached 91.28 EM and 91.64 F1 in the development set.

---

[1] https://pytorch-geometric.readthedocs.io
[2] https://github.com/studio-ousia/luke

Table 1. Hyper-parameters of our transformer-based module.

| Name | Value |
|---|---|
| Max Seq length | 512 |
| Max question length | 90 |
| Warmup ratio | 0.06 |
| Learning rate decay | linear |
| Weight decay | 0.01 |
| Adam $\beta_1$ | 0.9 |
| Adam $\beta_2$ | 0.98 |
| Adam $\epsilon$ | 1e-6 |

Table 2. Comparing the details of our configuration with the LUKE paper due to our limitations of computational resources.

| Name | Number of GPUs | Training epochs | Training time | Learning rate | Eval batch size | Dev Score F1 | Dev Score EM |
|---|---|---|---|---|---|---|---|
| LUKE [7] | 8 | 2 | 92min | 1e-5 | 32 | 91.4 | 90.8 |
| LUKE with Ours settings | 4 | 2 | 120min | 1e-5 | 32 | 90.96 | 90.4 |

### 4.3 Results

We compare our method (LUKE-Graph) to current state-of-the-art (SOTA) transformer-based and graph-based models on the ReCoRD dataset: BERT [1], Graph-BERT [27], SKG-BERT [11], KT-NET [12], XLNet-Verifier [33], KELM [9], RoBERTa [2], and LUKE [7]. We have selected only works related to our method and others are available on the leaderboard[3]. Their scores are taken directly from the leaderboard and literature [7,9]. As shown in Table 3, all the models are based on a single model.

Among transformer-based models, results on the dev set of ReCoRD show that the LUKE-Graph (ours) outperforms the former SOTA by +0.3 F1/+0.6 EM. This improvement demonstrates the effectiveness of adding a relational graph with attention heads and a question-aware gate on the LUKE model. As you can see from the results of transformer-based models, i.e., BERT, XLNET, RoBERTa, and LUKE, in Table 3, LUKE remarkably outperforms BERT-large by +19.2 F1/+19.3 EM and XLNET by +8.5 F1/+9.1 EM. While LUKE has lifted from RoBERTa and the only difference is adding entities to the inputs and using an entity-aware self-attention mechanism, it significantly achieves a gain of +0.6 F1/+0.6 EM over RoBERTa. Hence, due to the impressive results of the LUKE model, we choose it as our transformer-based model. However, it lacks a fair comparison of the transformer-based models, due to the computing and pretraining resources being different [34].

Among graph-based models, i.e., SKG-BERT, KT-NET, and KELM that use an external knowledge graph and BERT-large as a base model, the LUKE-Graph (ours) does not use any additional knowledge graph. However, experimental results demonstrate the LUKE-Graph offers a 18.7/19, 16.7/18.2, and 14.8/15 improvement in F1/EM over SKG-BERT, KT-NET, and KELM, respectively. Nevertheless, we note that the LUKE-Graph (ours) achieves the best performance compared to the other methods. To better visualization of the improvement process of the models based on the reported date in the leaderboard, we have illustrated the dev F1/EM results of the models in the chart forms in Fig. 6.

### 4.4 Ablation study

We perform several ablation studies on the ReCoRD development set to investigate the contribution of each module to the best model.

**Effect of graph module:** we investigate the impact of the stacking graph module on the LUKE model, as shown in Table 4. As we argued in the results section, the performance drops 0.4 on F1 and 0.55 on EM without the graph module. This proves the efficacy of taking into account the intuitive relations between entities on the performance of our method. Moreover, unlike other similar methods [13,23,35], we apply RGAT with a multi-head attention and a question-aware gating mechanism. However, the performance on F1/EM falls off 0.15/0.14, and 0.1/0.12 without the attention mechanism and without the question-aware gate in the graph network, respectively. Furthermore, we investigate the effect of removing the relational type in the graph and processing it by the RGAT module. If we treat all edges equally without distinguishing them by type (w/o relation type in Table 4), the F1/EM results degrade by 0.2/0.22, indicating that different information encoded by different types of edges is important to maintain good performance. We then inspect the impact of each type of relationship by removing each of them independently. Indeed, the nodes still exist in the graph and we only ablate corresponding edges between them. we either remove edges between those pairs of nodes that are in the same sentences (SENT-BASED edge), connections between those matching exactly (MATCH edge), or edges between the question placeholder and all other nodes (PLC edge). As you can see in Table 4, it seems that the placeholder connections (PLC edges) play a more major role compared to other connections. Because the missing entity (placeholder node) in question is supposed to be filled with candidate nodes and such an important connection is lost without the PLC edge.

---

[3] https://sheng-z.github.io/ReCoRD-explorer/

**Table 3.** Results on the ReCoRD dataset. All models are based on a single model. Note, missing results are indicated by "-". [±] Results reported in KELM paper, [+] Results reported in LUKE paper, [*] Results obtained by official LUKE code for a fair comparison, and others are taken from ReCoRD leaderboard.

| Name | Dev | | Test | |
|---|---|---|---|---|
| | F1 | EM | F1 | EM |
| Human | 91.64 | 91.28 | 91.69 | 91.31 |
| BERT-Base [1] | - | - | 56.1 | 54.0 |
| BERT-Large± [1] | 72.2 | 70.2 | 72.0 | 71.3 |
| Graph-BERT [27] | - | - | 63.0 | 60.8 |
| SKG-BERT± [11] | 71.6 | 70.9 | 72.8 | 72.2 |
| KT-NET± [12] | 73.6 | 71.6 | 74.8 | 73.0 |
| XLNet-Verifier+ [33] | 82.1 | 80.6 | 82.7 | 81.5 |
| KELM± [9] | 75.6 | 75.1 | 76.7 | 76.2 |
| RoBERTa+ [2] | 89.5 | 89.0 | 90.6 | 90.0 |
| LUKE+ [7] | 90.96* | 90.4* | 91.2 | 90.6 |
| LUKE-Graph (ours) | **91.36** | **90.95** | **91.5** | **91.2** |

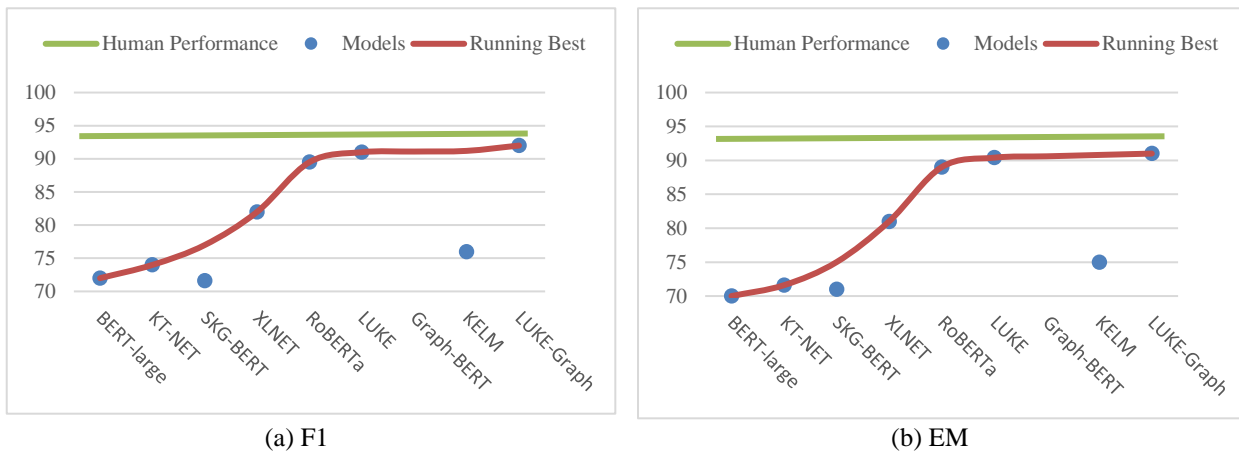

(a) F1    (b) EM

**Fig. 6.** Visualization of the improvement process of the models based on the reported date in the ReCoRD leaderboard for Dev F1 (a) and EM (b) results. The LUKE-Graph denotes our method.

**Impact of Different Graph Neural Networks:** we investigate the proposed model using the most popular GNN architectures in our graph module: RGAT [14], GATv2 [36], GAT [30], RGCN [31], and GCN [37]. Table 5 shows the effect of these architectures. All the experiments were performed using PyTorch Geometric. GCN is the most commonly used architecture in real-life applications that introduced an efficient layer-wise propagation rule for neural network models. However, it did not perform well in our experiments (-2.71 F1/-3.16 EM) due to not considering the importance between nodes and the type of connection between them. GAT extends GCN by considering a learnable attention mechanism to select the most relevant neighbors of each node. Therefore, GAT offers a 1.94 improvement in F1 over GCN in our model. GATv2 fixes the static attention problem of GAT by only modifying the order of attention operations in GAT. Hence, every node can attend to any other node in GATv2. In our experiment, GATv2 is more accurate than GAT and achieves a gain of +0.6 F1 over GAT. Additionally, Relational GCN (RGCN) and Relational GAT (RGAT) have been proposed as an extension of the previously mentioned models to the relational graph domain. With the difference that, in addition to considering the local relational structure, RGAT benefits from giving importance to different nodes dynamically and outperforms RGCN by +0.08 F1/+0.11 EM. Overall, the comparison of results in Table 5 demonstrates the power of RGAT in our method.

Table 4. Ablation results on the ReCoRD dev set. w/o stands for without.

| Model | Dev Results (%) | | | |
|---|---|---|---|---|
| | F1 | Δ | EM | Δ |
| Full Model | 91.36 | - | 90.95 | - |
| w/o graph module | 90.96 | 0.4 | 90.4 | 0.55 |
| w/o attention in the graph | 91.21 | 0.15 | 90.81 | 0.14 |
| w/o question-aware gate | 91.26 | 0.1 | 90.83 | 0.12 |
| w/o relational type | 91.16 | 0.2 | 90.73 | 0.22 |
| w/o SENT-BASED edges | 90.92 | 0.44 | 90.51 | 0.44 |
| w/o MATCH edges | 91.25 | 0.11 | 90.73 | 0.22 |
| w/o PLC edges | 90.72 | 0.64 | 90.18 | 0.77 |

Table 5. Impact of different graph neural networks.

| Model | Dev Results (%) | | | |
|---|---|---|---|---|
| | F1 | Δ | EM | Δ |
| LUKE+RGAT (ours) | 91.36 | - | 90.95 | - |
| +GATv2 [36] | 91.19 | 0.17 | 90.63 | 0.32 |
| +GAT [30] | 90.59 | 0.77 | 89.96 | 0.99 |
| +RGCN [31] | 91.28 | 0.08 | 90.84 | 0.11 |
| +GCN [37] | 88.65 | 2.71 | 87.79 | 3.16 |

## 5 Conclusion and Future Work

This paper presents the LUKE-Graph for MRC with commonsense reasoning, which leverages LUKE by considering the intuitive relationships between entities in long texts. Recent studies have integrated external knowledge and textual context into a unified data structure for processing commonsense reasoning. However, there is a challenge when it comes to selecting the best subgraph and the most relevant entity in KGs, especially for ambiguous words. Here, the LUKE-Graph differs from these studies; it converts the entities in the document and their relations into a heterogeneous graph. We also use RGAT to improve the entity representation encoded by LUKE using graph information. Furthermore, we optimize RGAT to Gated-RGAT by incorporating the question information during reasoning through a gating mechanism. The enrichment of the contextual representation of the pre-trained LUKE with Gated-RGAT proved valuable for the ReCoRD dataset and resulted in a 0.6% improvement of EM over the baseline LUKE.

As Future works, we will verify our model on more challenging datasets with multi-hop reasoning problems, and we would like to incorporate the importance of the relationship between entities in some way rather than using a separate graph module.